\documentclass[preprint,12pt]{elsarticle}

\usepackage{footnote}
\makesavenoteenv{tabular}

\usepackage{soul}
\sethlcolor{orange}  

\usepackage{amsthm}
\usepackage{float}
\usepackage{amsmath}
\DeclareMathOperator{\pred}{Pred}
\DeclareMathOperator{\layerv}{Layer^v}
\DeclareMathOperator{\layerf}{Layer^f}
\DeclareMathOperator{\messageFN}{Message}
\DeclareMathOperator{\updateFN}{Update}
\DeclareMathOperator{\aggregateFN}{Aggregate}
\DeclareMathOperator{\attendFN}{Attend}

\usepackage{makecell} 

\usepackage[dvipsnames]{xcolor}
\usepackage{graphicx}
\usepackage{epstopdf}
\usepackage{float}
\usepackage{bm,upgreek}
\usepackage{amsfonts}
\usepackage{enumitem}
\usepackage{mathtools}
\usepackage{multirow}
\usepackage{booktabs}
\usepackage[subnum]{cases}
\usepackage{setspace}
\usepackage{color}
\usepackage{blkarray, bigstrut} 
\usepackage[caption=false,labelformat=simple]{subfig}

\usepackage{xcolor}
\usepackage{pgfplots}
\pgfplotsset{compat=1.5,
  tick label style={font=\small},
  label style={font=\small}}
\usepackage{pgfplotstable}
\usepgfplotslibrary{statistics}
\usepackage{tikz}

\usepackage[european]{circuitikz}
\newcommand{\bushere}[2]{%
    coordinate(tmp)
    ++(0,1) node[above]{#1} edge[ultra thick] ++(0,-2)
    ++(0,-2) node[below]{#2}
    (tmp)
}

\usepackage{pgfplots}
\usepackage{pgfplotstable}
\pgfplotsset{compat=1.7}
\usepgfplotslibrary{groupplots}

\usepackage[toc,acronym]{glossaries}
\usepackage{mathtools,nccmath}
\usepackage[ruled]{algorithm}
\usepackage{algpseudocode,algorithmicx}

\algnewcommand{\LineComment}[1]{\State \(\#\) #1}

\allowdisplaybreaks

\pgfplotsset{legend image with text/.style={legend image code/.code={%
\node[anchor=west, align=right] at (0.0cm,0cm) {#1};}},}

\makeatletter
\def\myline{\pgfutil@ifnextchar[{\my@line}{\my@line[]}}%
\def\my@line[#1](#2)(#3){%
\tikz[overlay] \draw[#1]  (#2)--(#3); 
}%
\algrenewcommand\algorithmicindent{1.0em}%

\makeatletter
\renewcommand{\ALG@beginalgorithmic}{\small}
\makeatother



\usepackage{etoolbox}
\BeforeBeginEnvironment{algorithmic}{\kern 0.3\baselineskip}
\AfterEndEnvironment{algorithmic}{\kern -0.1\baselineskip}

\makeatletter
\renewcommand{\ALG@beginalgorithmic}{\footnotesize}
\makeatother

\pgfplotsset{
    box plot/.style={
        /pgfplots/.cd,
        black,
        only marks,
        mark=-,
        mark size=\pgfkeysvalueof{/pgfplots/box plot width},
        /pgfplots/error bars/y dir=plus,
        /pgfplots/error bars/y explicit,
        /pgfplots/table/x index=\pgfkeysvalueof{/pgfplots/box plot x index},
    },
    box plot box/.style={
        /pgfplots/error bars/draw error bar/.code 2 args={%
            \draw  ##1 -- ++(\pgfkeysvalueof{/pgfplots/box plot width},0pt) |- ##2 -- ++(-\pgfkeysvalueof{/pgfplots/box plot width},0pt) |- ##1 -- cycle;
        },
        /pgfplots/table/.cd,
        y index=\pgfkeysvalueof{/pgfplots/box plot box top index},
        y error expr={
            \thisrowno{\pgfkeysvalueof{/pgfplots/box plot box bottom index}}
            - \thisrowno{\pgfkeysvalueof{/pgfplots/box plot box top index}}
        },
        /pgfplots/box plot
    },
    box plot top whisker/.style={
        /pgfplots/error bars/draw error bar/.code 2 args={%
            \pgfkeysgetvalue{/pgfplots/error bars/error mark}%
            {\pgfplotserrorbarsmark}%
            \pgfkeysgetvalue{/pgfplots/error bars/error mark options}%
            {\pgfplotserrorbarsmarkopts}%
            \path ##1 -- ##2;
        },
        /pgfplots/table/.cd,
        y index=\pgfkeysvalueof{/pgfplots/box plot whisker top index},
        y error expr={
            \thisrowno{\pgfkeysvalueof{/pgfplots/box plot box top index}}
            - \thisrowno{\pgfkeysvalueof{/pgfplots/box plot whisker top index}}
        },
        /pgfplots/box plot
    },
    box plot bottom whisker/.style={
        /pgfplots/error bars/draw error bar/.code 2 args={%
            \pgfkeysgetvalue{/pgfplots/error bars/error mark}%
            {\pgfplotserrorbarsmark}%
            \pgfkeysgetvalue{/pgfplots/error bars/error mark options}%
            {\pgfplotserrorbarsmarkopts}%
            \path ##1 -- ##2;
        },
        /pgfplots/table/.cd,
        y index=\pgfkeysvalueof{/pgfplots/box plot whisker bottom index},
        y error expr={
            \thisrowno{\pgfkeysvalueof{/pgfplots/box plot box bottom index}}
            - \thisrowno{\pgfkeysvalueof{/pgfplots/box plot whisker bottom index}}
        },
        /pgfplots/box plot
    },
    box plot median/.style={
        /pgfplots/box plot,
        /pgfplots/table/y index=\pgfkeysvalueof{/pgfplots/box plot median index},
        semithick,black
    },
    box plot width/.initial=1em,
    box plot x index/.initial=0,
    box plot median index/.initial=1,
    box plot box top index/.initial=2,
    box plot box bottom index/.initial=3,
    box plot whisker top index/.initial=4,
    box plot whisker bottom index/.initial=5,
}

\algnewcommand\algorithmicforeach{\textbf{for each}}
\algdef{S}[FOR]{ForEach}[1]{\algorithmicforeach\ #1\ \algorithmicdo}

\pgfplotsset{grid style={dotted,gray}}

\usepackage{amssymb}

\journal{Sustainable Energy, Grids and Networks}

\begin{document}

\begin{frontmatter}



\title{Graph Neural Networks on Factor Graphs for Robust, Fast, and Scalable Linear State Estimation with PMUs}


\author[inst1]{Ognjen Kundacina}
\author[inst2]{Mirsad Cosovic}
\author[inst1]{Dragisa Miskovic}
\author[inst3]{Dejan Vukobratovic}

\affiliation[inst1]{organization={The Institute for Artificial Intelligence Research and Development of Serbia},
            city={Novi Sad},
            postcode={21000}, 
            country={Serbia}}
            
\affiliation[inst2]{organization={Faculty of Electrical Engineering},
            addressline={University of Sarajevo}, 
            city={Sarajevo},
            postcode={71000}, 
            country={Bosnia and Herzegovina}}

\affiliation[inst3]{organization={Faculty of Technical Sciences},
            addressline={University of Novi Sad}, 
            city={Novi Sad},
            postcode={21000}, 
            country={Serbia}}

\begin{abstract}
As phasor measurement units (PMUs) become more widely used in transmission power systems, a fast state estimation (SE) algorithm that can take advantage of their high sample rates is needed. To accomplish this, we present a method that uses graph neural networks (GNNs) to learn complex bus voltage estimates from PMU voltage and current measurements. We propose an original implementation of GNNs over the power system's factor graph to simplify the integration of various types and quantities of measurements on power system buses and branches. Furthermore, we augment the factor graph to improve the robustness of GNN predictions. This model is highly efficient and scalable, as its computational complexity is linear with respect to the number of nodes in the power system. Training and test examples were generated by randomly sampling sets of power system measurements and annotated with the exact solutions of linear SE with PMUs. The numerical results demonstrate that the GNN model provides an accurate approximation of the SE solutions. Furthermore, errors caused by PMU malfunctions or communication failures that would normally make the SE problem unobservable have a local effect and do not deteriorate the results in the rest of the power system.
\end{abstract}



\begin{keyword}
State Estimation, Graph Neural Networks, Machine Learning, Power Systems, Real Time Systems
\end{keyword}

\end{frontmatter}


\section{Introduction}
\label{sec:sample1}

\textbf{Motivation:} The state estimation (SE), which estimates the set of power system state variables based on the available set of measurements, is an essential tool used for the power system's monitoring and operation \cite{monticelli2000SE}. The increasing deployment of phasor measurement units (PMUs) in transmission power systems calls for a fast state estimator to take full advantage of their high sample rates. Solving PMU-only SE, which is represented as a linear weighted least-squares (WLS) problem, involves matrix factorisations, which can be difficult in cases where the matrix is ill-conditioned. It is common practice to neglect the phasor measurement covariances represented in rectangular coordinates \cite{exposito}. This can make the SE problem much easier to solve, but it also results in a computational complexity of nearly $\mathcal{O}(n^2)$ for sparse matrices, where $n$ is the number of power system buses. It is challenging to use this approach for real-time monitoring of large power systems. Recent advancements in graph neural networks (GNNs) \cite{pmlr-v70-gilmer17a, GraphRepresentationLearningBook, velickovic2018graph} open up novel possibilities for developing power system algorithms with linear computational complexity and potential distributed implementation. One of the key benefits of using GNNs in power systems instead of conventional deep learning methods is that the prediction model is not restricted to training and test examples of fixed power system topologies. GNNs take advantage of the graph structure of the input data, and hence are permutation invariant, have fewer trainable parameters, and require less storage space. Furthermore, the inference phase of many GNN models can be distributed and performed locally, since the prediction of the state variable of a node only requires measurements from the local neighbourhood, which extends to a specific distance from the node being predicted.
GNNs, as a model-free approach, can be useful when power system parameters are uncertain or unreliable \cite{antona2018PSUncertainty}. However, if necessary, GNNs can incorporate power system parameters through the use of edge input features \cite{gong2019EdgeFeatures}.

\textbf{Literature Review:} 
The popularity of deep learning in the field of power systems has been well-documented in recent research \cite{LOPEZ2020AnnMicrogrids, PANG2022100660, FELLNER2022100851}, with several studies showing that it can be used to learn the solutions to computationally intensive power system analysis algorithms. In \cite{zhang2019}, the authors used a combination of recurrent and feed-forward neural networks to solve the power system SE problem using measurement data and the history of network voltages. Another study, \cite{zamzam2019}, provides an example of training a feed-forward neural network to initialise the network voltages for a Gauss-Newton distribution system SE solver.

GNNs are starting to gain widespread use in power systems, providing solutions to a wide range of prediction tasks, including fault location \cite{chen2020FaultLocation}, stability assessment \cite{ZHANG2022Stability}, power system parameter identification \cite{XIA2022ParamIdentification}, and detecting false data injection attacks \cite{HAN2023FalseData}. GNNs are also being used for control and optimisation tasks, such as optimal scheduling \cite{XING2023Scheduling} and volt-var control \cite{LEE2022VVC}. Several studies have proposed the use of GNNs for power flow problems, including \cite{donon2019graphneuralsolver} and \cite{bolz2019PFapproximator}. In these works, power flows in the system are predicted based on power injection data labelled by a traditional power flow solver. Similarly, \cite{wang2020ProbabPF} suggests using a trained GNN as an alternative to computationally expensive probabilistic power flow methods, which calculate probability density functions of unknown variables. Different approaches propose training a GNN in an unsupervised manner to perform power flow calculations by minimising the violation of Kirchhoff's law \cite{donon2020graphneuralsolverJOURNAL} or power balance error \cite{LOPEZ2023GNNpowerFlow} at each bus, thus avoiding the need for labelled data from a conventional power flow solver. 

In \cite{pagnier2021physicsinformed}, the authors propose a combined model- and data-based approach using GNNs for power system parameter and state estimation. The model predicts power injections and consumptions in nodes where voltage and phase measurements are taken, but it does not consider branch measurements and other types of node measurements in its calculations. In \cite{Yang2022RobustPSSEDataDrivenPriors}, the authors train a GNN by propagating simulated or measured voltages through the graph to learn the voltage labels from a historical dataset, and then use the GNN as a regularisation term in the SE loss function. However, the proposed GNN only uses node voltage measurements and does not consider other types of measurements, although they are handled in other parts of the algorithm. Another feed-forward neural network learns the solutions that minimise the SE loss function, resulting in an acceleration of the SE solution with $\mathcal{O}(n^2)$ computational complexity at inference time. In \cite{TGCN_SE_2021}, state variables are predicted based on a time-series of node voltage measurements, and the authors solve the SE problem using GNNs with gated recurrent units.

\textbf{Contributions:} This paper proposes using a specialised GNN model for linear SE with only PMUs in positive sequence power transmission systems. To provide fast and accurate predictions during the evaluation phase, the GNN is trained using the inputs and solutions from a traditional linear SE solver. The following are the paper's main contributions:
\begin{itemize}[leftmargin=*]
\item Inspired by \cite{Satorras2021NeuralEB}, we present the first use of GNNs on factor graphs \cite{Kschischang2001FactorGraphs} for the SE problem, instead of using the bus-branch power system model. This enables trivial integration and exclusion of any type and number of measurements on the power system buses and branches, by adding or removing the corresponding nodes in the factor graph. Furthermore, the factor graph is augmented by adding direct connections between variable nodes that are $2^{nd}$-order neighbours to improve information propagation during neighbourhood aggregation, particularly in unobservable scenarios when the loss of the measurement data occurs.
\item We present a graph attention network (GAT) \cite{velickovic2018graph} model, with the architecture customised for the proposed heterogeneous augmented factor graph, to solve the SE problem. GNN layers that aggregate into factor and variable nodes have separate sets of trainable parameters. Furthermore, separate sets of parameters are used for variable-to-variable and factor-to-variable message functions in GNN layers that aggregate into variable nodes.
\item Given the sparsity of the power system's graph, and the fact that node degree does not increase with the total number of nodes, the proposed approach has $\mathcal{O}(n)$ computational complexity, making it suitable for large-scale power systems. The inference of the trained GNN is easy to distribute and parallelise. Even in the case of centralised SE implementation, the processing can be done using distributed computation resources, such as graphical-processing units.
\item We demonstrate that the number of trainable parameters in the proposed GNN-based SE model is constant, while it grows quadratically with the number of measurements in conventional deep learning approaches.
\item We study the local-processing nature of the proposed model and show that significant degradation of results caused by PMU or communication failures affects only the local neighbourhood of the node where the failure occurred.
\end{itemize}

We note that this paper significantly extends the conference version \cite{kundacina2022state}. The GNN model architecture is improved and described in greater detail, and the algorithm's performance and scalability are enhanced by using binary encoding for variable node indices. Also, GNN inputs now additionally include measurement covariances represented in rectangular coordinates, and discussions on computational complexity, distributed implementation, and robustness to outliers are included. We expand the numerical results section with experiments that include more cases of measurement variances and redundancies and evaluate the algorithm's scalability on larger power systems with sizes ranging from 30 to 2000 buses. Finally, we include a detailed comparison of the proposed GNN model with the state-of-the-art deep learning-based a SE approach \cite{zhang2019}, as well as the approximate linear WLS solution of SE with PMUs \cite{exposito}, which neglects covariances of measurement represented in rectangular coordinates. The latter is a well-established approach that is widely used in the field of power system SE, based on solely PMU inputs.

The rest of the paper is organised as follows. Section 2 provides a formulation of the linear SE with PMUs problem. Section 3 gives the theoretical foundations of GNNs and defines the augmented factor graph and describes the GNN architecture. Section 4 shows numerical results from various test scenarios, while Section 5 draws the main conclusions of the work.

\section{Linear State Estimation with PMUs}
\label{sec:lse}

The SE algorithm estimates the values of the state variables $\mathbf{x}$ based on the knowledge of the network topology and parameters, and measured values obtained from the measurement devices spread across the power system. 

The power system network topology is described by the bus-branch model and can be represented using a graph $\mathcal{G} =(\mathcal{H},\mathcal{E})$, where the set of nodes $\mathcal{H} = \{1,\dots,n \}$ represents the set of buses, while the set of edges $\mathcal{E} \subseteq \mathcal{H} \times \mathcal{H}$ represents the set of branches of the power network. The branches of the network are defined using the two-port $\pi$-model. More precisely, the branch $(i,j) \in \mathcal{E}$ between buses $\{i,j\} \in \mathcal{H}$ can be modelled using complex expressions:
\begin{equation}
  \begin{bmatrix}
    {I}_{ij} \\ {I}_{ji}
  \end{bmatrix} =
  \begin{bmatrix}
    \cfrac{1}{\tau_{ij}^2}(y_{ij} + y_{\text{s}ij}) & -\alpha_{ij}^*{y}_{ij}\\
    -\alpha_{ij}{y}_{ij} & {y}_{ij} + y_{\text{s}ij}
  \end{bmatrix}  
  \begin{bmatrix}
    {V}_{i} \\ {V}_{j}
  \end{bmatrix},
  \label{unified}
\end{equation} 
where the parameter $y_{ij} = g_{ij} + \text{j}b_{ij}$ represents the branch series admittance, half of the total branch shunt admittance (i.e., charging admittance) is given as $y_{\text{s}ij} = \text{j}b_{si}$. Further, the transformer complex ratio is defined as $\alpha_{ij} = (1/\tau_{ij})\text{e}^{-\text{j}\phi_{ij}}$, where $\tau_{ij}$ is the transformer tap ratio magnitude, while $\phi_{ij}$ is the transformer phase shift angle. It is important to remember that the transformer is always located at the bus $i$ of the branch described by \eqref{unified}. Using the branch model defined by \eqref{unified}, if $\tau_{ij} = 1$ and $\phi_{ij} = 0$ the system of equations describes a line. In-phase transformers are defined if $\phi_{ij} = 0$ and $y_{\text{s}ij} = 0$, while phase-shifting transformers are obtained if $y_{\text{s}ij} = 0$. The complex expressions ${I}_{ij}$ and ${I}_{ji}$ define branch currents from the bus $i$ to the bus $j$, and from the bus $j$ to the bus $i$, respectively. The complex bus voltages at buses $\{i,j\}$ are given as $V_i$ and $V_j$, respectively.  

PMUs measure complex bus voltages and complex branch currents. More precisely, phasor measurement provided by PMU is formed by a magnitude, equal to the root-mean-square value of the signal, and phase angle \cite[Sec.~5.6]{phadke}. The PMU placed at the bus measures bus voltage phasor and current phasors along all branches incident to the bus \cite{exposito}. Thus, the PMU outputs phasor measurements in polar coordinates. In addition, PMU outputs can be observed in the rectangular coordinates with real and imaginary parts of the bus voltage and branch current phasors. In that case, the vector of state variables $\mathbf{x}$ can be given in rectangular coordinates $\mathbf x \equiv[\mathbf{V}_\mathrm{re},\mathbf{V}_\mathrm{im}]^{\mathrm{T}}$, where we can observe real and imaginary components of bus voltages as state variables:   
\begin{equation}
   \begin{aligned}
    \mathbf{V}_\mathrm{re}&=\big[\Re({V}_1),\dots,\Re({V}_n)\big]\\
	\mathbf{V}_\mathrm{im}&=\big[\Im({V}_1),\dots,\Im({V}_n)\big].     
   \end{aligned}
   \label{rect_coord}
\end{equation} 

Using rectangular coordinates, we obtain the linear system of equations defined by voltage and current measurements obtained from PMUs. The measurement functions corresponding to the bus voltage phasor measurement on the bus $i \in \mathcal{H}$ are simply equal to: 
\begin{equation}
    \begin{aligned}
        f_{\Re\{V_i\}}(\cdot) = \Re\{V_i\}\\
        f_{\Im\{V_i\}}(\cdot) = \Im\{V_i\}.
    \end{aligned}  
    \label{measFunction1}
\end{equation}
According to the unified branch model \eqref{unified}, functions corresponding to the branch current phasor measurement vary depending on where the PMU is located. If PMU is placed at the bus $i$, functions are given as:
\begin{equation}
    \begin{aligned}
        f_{\Re\{I_{ij}\}}(\cdot) &= q \Re\{V_{i}\} - w \Im\{V_{i}\} - (r-t) \Re\{V_{j}\} + (u+p) \Im\{V_{j}\} \\
        f_{\Im\{I_{ij}\}}(\cdot) &= w \Re\{V_{i}\} + q \Im\{V_{i}\} - (u+p) \Re\{V_{j}\} - (r-t) \Im\{V_{j}\},
    \end{aligned}
    \label{measFunction2}
\end{equation}
where $q =$ $g_{ij}/\tau_{ij}^2$, $w =$ $(b_{ij} + b_{si})/\tau_{ij}^2$, $r =$ $(g_{ij}/\tau_{ij})$ $\cos\phi_{ij}$, $t =$ $(b_{ij}/\tau_{ij})$ $\sin\phi_{ij}$, $u =$ $(b_{ij}/\tau_{ij})$ $\cos\phi_{ij}$, $p =$ $(g_{ij}/\tau_{ij})$ $\sin\phi_{ij}$. In the case where PMU is installed at the bus $j$, measurement functions are: 
\begin{equation}
    \begin{aligned}
        f_{\Re\{I_{ji}\}}(\cdot) &= z \Re\{V_{j}\} - e \Im\{V_{j}\} - (r+t) \Re\{V_{i}\} + (u-p) \Im\{V_{i}\} \\
        f_{\Im\{I_{ji}\}}(\cdot) &= e \Re\{V_{j}\} + z \Im\{V_{j}\} - (u-p) \Re\{V_{i}\} - (r+t) \Im\{V_{i}\},
    \end{aligned}
    \label{measFunction3}
\end{equation}
where $z = g_{ij}$ and $e =$ $b_{ij} + b_{si}$. The presented model represents the system of linear equations, where the solution can be found by solving the linear weighted least-squares problem: 
\begin{equation}
    \left(\mathbf H^{T} \mathbf \Sigma^{-1} \mathbf H \right) \mathbf x =
		\mathbf H^{T} \mathbf \Sigma^{-1} \mathbf z,    
	\label{SE_system_of_lin_eq}
\end{equation}
where the Jacobian matrix $\mathbf {H} \in \mathbb {R}^{2m \times 2n}$ is defined according to measurement functions \eqref{measFunction1}-\eqref{measFunction3}, $2m$ is the total number of linear equations, where $m$ is the number of phasor measurements. The observation error covariance matrix is given as $\mathbf {\Sigma} \in \mathbb {R}^{2m \times 2m}$, and the vector $\mathbf z \in \mathbb {R}^{2m}$ contains measurement values given in rectangular coordinate system.

The main disadvantage of this approach is that measurement errors correspond to polar coordinates (i.e., magnitude and angle errors), whereas the covariance matrix must be transformed from polar to rectangular coordinates \cite{zhou2016phasorsMeasSE}. As a result, measurement errors are correlated and the covariance matrix $\mathbf {\Sigma}$ does not have a diagonal form. Despite that, because of the lower computational effort, the non-diagonal elements of the covariance matrix $\mathbf {\Sigma}$ are usually neglected, which has an effect on the accuracy of the state estimation \cite{exposito}. Using the classical theory of propagation of uncertainty, the variance in the rectangular coordinate system can be obtained using variances in the polar coordinate system. For example, let us observe the voltage phasor measurement at the bus $i$, where PMU outputs the voltage magnitude measurement value $z_{|V_i|}$ with corresponding variance $v_{|V_i|}$, and voltage phase angle measurement $z_{\theta_i}$ with variance $v_{\theta_i}$. Then, variances in the rectangular coordinate system can be obtained as:
\begin{equation}
    \begin{aligned}
        v_{\Re\{V_i\}} &= v_{|V_i|} (\cos z_{\theta_i})^2 + v_{\theta_i} (z_{|V_i|} \sin z_{\theta_i})^2\\
        v_{\Im\{V_i\}} &= v_{|V_i|} (\sin z_{\theta_i})^2 + v_{\theta_i} (z_{|V_i|} \cos z_{\theta_i})^2.
    \end{aligned}
\end{equation}
Analogously, we can easily compute variances related to current measurements $v_{\Re\{I_{ij}\}}$, $v_{\Im\{I_{ij}\}}$ or $v_{\Re\{I_{ji}\}}$, $v_{\Im\{I_{ji}\}}$. 
We will refer to the solution of \eqref{SE_system_of_lin_eq} in which measurement error covariances are neglected to avoid the computationally demanding inversion of the non-diagonal matrix $\mathbf {\Sigma}$ as an \textit{approximative WLS SE solution} \cite{exposito}. 

In this work, we will investigate if the GNN model trained with measurement values, variances, and covariances labelled with the exact solutions of \eqref{SE_system_of_lin_eq} is more accurate than the approximative WLS SE, which neglects the covariances. Inference performed using the trained GNN model scales linearly with the number of power system buses, making it significantly faster than both the approximate and the exact solver of \eqref{SE_system_of_lin_eq}.

\section{Graph Neural Network-based State Estimation}
In this section, we present the GNN theory foundations, augmentation techniques for the power system's factor graph, the details of the proposed GNN architecture, and analyse the computational complexity and the distributed implementation of the GNN model's inference.

\subsection{Basics of Spatial Graph Neural Networks} 

\begin{figure}[!t]
\centering

\begin{tikzpicture} [scale=0.71, transform shape]

\tikzset{
    box/.style={draw, fill=Goldenrod, minimum width=1.5cm, minimum height=0.8cm}}
    
\begin{scope}[local bounding box=graph]

\node [box]  (message1) at (-0.5, 1.5) {Message};
\node [box]  (message2) at (-0.5, 0) {Message};
\node [box]  (gat) at (3, 0) {$\aggregateFN$};
\node [box]  (update) at (6.3, 0) {$\updateFN$};


\draw[-stealth] (-1.8, 0) -- (message2.west) node[at start,left]{$\mathbf{h_{2}}^{k-1}; \mathbf{h_{j}}^{k-1}$};

\draw[-stealth] (-1.8, 1.5) -- (message1.west) node[at start,left]{$\mathbf{h_{1}}^{k-1}; \mathbf{h_{j}}^{k-1}$};

\draw[-stealth] (message1.east) -- (gat.west) node[near start,right]{$\mathbf{m_{1,j}}^{k-1}$};

\draw[-stealth] (message2.east) -- (gat.west) node[midway,below]{$\mathbf{m_{2,j}}^{k-1}$};
	
\draw[-stealth] (1.9, -1) -- (gat.west)
	node[at start,left]{$...$};
	
\draw[-stealth] (gat.east) -- (update.west) node[midway,above]{$\mathbf {m_j}^{k-1}$};

\draw[-stealth] (update.east) -- (7.6, 0) node[at end,right]{$\mathbf{h_{j}}^{k}$};

\draw[-stealth] (6.3, -1.0) -- (update.south) node[at start,below]{$\mathbf{h_{j}}^{k-1}$};

\end{scope}

\end{tikzpicture}

\caption{The detailed structure of a single GNN layer, where functions with trainable parameters are highlighted in yellow.}
    \label{GNNlayerDetails}
\end{figure}
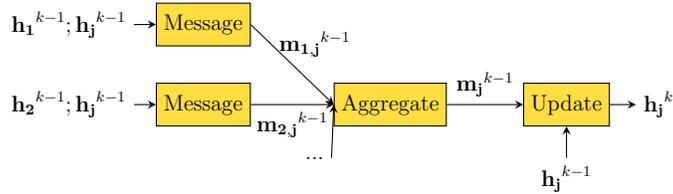

The spatial GNNs used in this study learn over graph-structured data using a recursive neighbourhood aggregation scheme, also known as the message passing procedure \cite{pmlr-v70-gilmer17a}. This results in a $s$-dimensional vector embedding $\mathbf h \in \mathbb {R}^{s}$ of each node, which captures the information about the node's position in the graph, as well as its own and the input features of the neighbouring nodes. The GNN layer, which implements one iteration of the recursive neighbourhood aggregation consists of several functions, that can be represented using a trainable set of parameters, usually in the form of the feed-forward neural networks. One of the functions is the message function $\messageFN(\cdot|\theta^{\messageFN}): \mathbb {R}^{2s} \mapsto \mathbb {R}^{u}$, which outputs the message $\mathbf m_{i,j} \in \mathbb {R}^{u}$ between two node embeddings, $\mathbf h_i$ and $\mathbf h_j$. The second one is the aggregation function $\aggregateFN(\cdot|\theta^{\aggregateFN}): \mathbb {R}^{\textrm{deg}(j) \cdot u} \mapsto \mathbb {R}^{u}$, which defines the way in which the incoming neighbouring messages are combined and outputs the aggregated messages, denoted as $\mathbf {m_j} \in \mathbb {R}^{u}$ for node $j$. The output of one iteration of neighbourhood aggregation process is the updated node embedding obtained by applying the update function $\updateFN(\cdot|\theta^{\updateFN}): \mathbb {R}^{u} \mapsto \mathbb {R}^{s}$ on the aggregated messages. The recursive neighbourhood aggregation process is repeated a predefined number of iterations $K$, also known as the number of GNN layers. When executing the recursive neighbourhood aggregation process for a single node, GNN collects messages from a set of nodes that extends up to $K$ edges away from the node. This set of nodes is referred to as the node's $K$-hop neighbourhood throughout this work. The initial node embedding values are equal to the $l$-dimensional node input features, linearly transformed to the initial node embedding $\mathbf {h_j}^0 \in \mathbb {R}^{s}$. The superscript of node embeddings and calculated messages indicates the iteration they belong to. One iteration of the neighbourhood aggregation process for the $k^{th}$ GNN layer is depicted in Fig.~\ref{GNNlayerDetails} and also described by equations \eqref{gnn_equations}:
\begin{equation}
    \begin{gathered}
        \mathbf {m_{i,j}}^{k-1} = \messageFN( \mathbf {h_i}^{k-1}, \mathbf {h_j}^{k-1} | \theta^{\messageFN})\\
        \mathbf {m_j}^{k-1} = \aggregateFN(\{{\mathbf m_{i,j}}^{k-1} | i \in \mathcal{N}_j\} | \theta^{\aggregateFN})\\
        \mathbf {h_j}^k = \updateFN(\mathbf {m_j}^{k-1}, \mathbf {h_j}^{k-1}) | \theta^{\updateFN})\\
        k \in \{1,\dots,K\},
    \end{gathered}
    \label{gnn_equations}
\end{equation}
where $\mathcal{N}_j$ denotes the $1$-hop neighbourhood of the node $j$. The output of the whole process are final node embeddings $\mathbf {h_j}^{K}$ which can be used for the classification or regression over the nodes, edges, or the whole graph. In the case of node-level regression, the final embeddings are passed through the additional nonlinear function, creating the predictions of the GNN model for the set of inputs fed into the nodes and their neighbours. The GNN model is trained by backpropagation of the loss function between the predictions and the ground-truth values over the whole computational graph. We refer the reader to \cite{GraphRepresentationLearningBook} for a more detailed introduction to graph representation learning and GNNs.

\subsection {Graph Attention Networks}
One important decision when creating a GNN model is selecting the aggregation function. Common choices include sum, mean, min and max pooling, and graph convolution \cite{KipfW17_GCN}. One common drawback of these approaches is weighting incoming messages from all neighbours equally, or using weights calculated from the structural properties of the graph (e.g., node degrees). Graph attention networks (GATs) \cite{velickovic2018graph} propose using the attention-based aggregation, where the weights are equal to each neighbour's importance score, which is learned from their embeddings.
 
The attention mechanism introduces an additional set of trainable parameters in the aggregation function, usually implemented as a feed-forward neural network $\attendFN(\cdot|\theta^{\attendFN}): \mathbb {R}^{2s} \mapsto \mathbb {R}$ applied over the concatenated embeddings of the node $j$ and each of its neighbours:
 
\begin{equation}
    e_{i,j} = \attendFN(\mathbf {h_i}, \mathbf {h_j} | \theta^{\attendFN}).
    \label{unnormalized_attention}
\end{equation}

The final attention weights $a_{i,j} \in \mathbb {R}$ are calculated by normalising the output $e_{i,j} \in \mathbb {R}$ using the softmax function:
 \begin{equation}
    a_{i,j} = \frac{exp({e_{i,j}})}{\sum_{i' \in \mathcal{N}_j} exp({e_{i',j}})}.
    \label{normalized_attention}
\end{equation}

\subsection{Power System Factor Graph Augmentation and the Proposed GNN Architecture} 
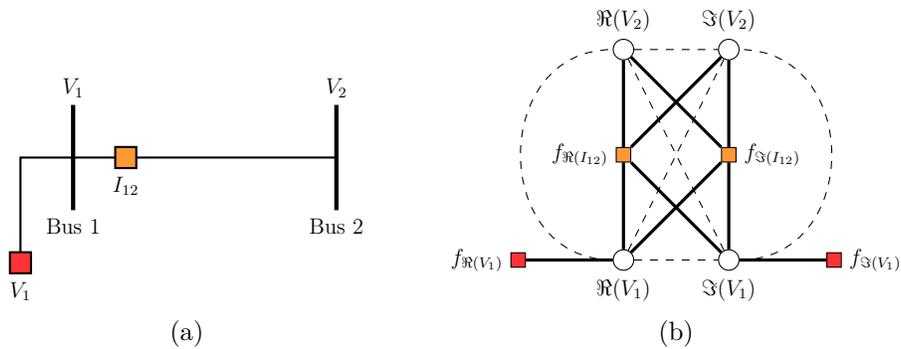
\begin{figure}[htbp]
    \centering
    \subfloat[]{
    \centering
    \begin{tikzpicture}[thick, scale=0.7, transform shape]
        \tikzset{
            factorVoltage/.style={draw=black,fill=red!80, minimum size=4mm},
            factorCurrent/.style={draw=black,fill=orange!80, minimum size=4mm}}
        \draw (0,0) 
            node[factorVoltage, label=below:$V_{1}$](L11){} 
            -- ++(0,2)
            -- ++(1,0)
            \bushere{${V}_1$}{Bus 1}
            -- ++(1,0)
            node[factorCurrent, label=below:$I_{12}$](L2){} 
            to[] ++(4,0)
            \bushere{${V}_2$}{Bus 2}
            ;
    \end{tikzpicture}
    }
    \hfil
    \subfloat[]{
    \begin{tikzpicture} [scale=0.7, transform shape]
        \tikzset{
            varNode/.style={circle,minimum size=2mm,fill=white,draw=black},
            factorVoltage/.style={draw=black,fill=red!80, minimum size=2mm},
            factorCurrent/.style={draw=black,fill=orange!80, minimum size=2mm},
            edge/.style={very thick,black},
            edge2/.style={dashed,black}}
        \begin{scope}[local bounding box=graph]
            \node[factorVoltage, label=left:$f_{\Re({V}_1)}$] (f1) at (-3, 1 * 2) {};
            \node[factorVoltage, label=right:$f_{\Im({V}_1)}$] (f4) at (3, 1 * 2) {};
            \node[varNode, label=below:$\Re({V}_1)$] (v1) at (-1, 1 * 2) {};
            \node[varNode, label=below:$\Im({V}_1)$] (v3) at (1, 1 * 2) {};
            \node[factorCurrent, label=left:$f_{\Re({I}_{12})}$] (f2) at (-1, 2 * 2) {};
            \node[factorCurrent, label=right:$f_{\Im({I}_{12})}$] (f5) at (1, 2 * 2) {};
            \node[varNode, label=above:$\Re({V}_2)$] (v2) at (-1, 3 * 2) {};
            \node[varNode, label=above:$\Im({V}_2)$] (v4) at (1, 3 * 2) {};
            
            \draw[edge] (f1) -- (v1);
            \draw[edge] (f4) -- (v3);
            \draw[edge] (f2) -- (v1);
            \draw[edge] (f5) -- (v3);
            \draw[edge] (f5) -- (v1);
            \draw[edge] (f2) -- (v3);
            \draw[edge] (f2) -- (v2);
            \draw[edge] (f2) -- (v4);
            \draw[edge] (f5) -- (v2);
            \draw[edge] (f5) -- (v4);
            \draw[edge2] (v1) to [out=180,in=180,looseness=1.5] (v2);
            \draw[edge2] (v1) -- (v3);
            \draw[edge2] (v1) -- (v4);
            \draw[edge2] (v2) -- (v3);
            \draw[edge2] (v2) -- (v4);
            \draw[edge2] (v3) to [out=0,in=0,looseness=1.5] (v4);
            
        \end{scope}
    \end{tikzpicture}
    }
    \caption{Subfigure (a) shows a simple two-bus power system, and Subfigure (b) displays the corresponding factor graph (full-line edges) and augmented factor graph (all edges). Variable nodes are represented as circles, and factor nodes are depicted as squares, coloured differently to distinguish between measurement types.}
    \label{toyFactorGraph}
    
\end{figure}

Inspired by recent work on using probabilistic graphical models for power system SE \cite{cosovic2019bpse}, we first create a GNN over a graph with a factor graph topology. This bipartite graph consists of factor and variable nodes, and edges between them. Variable nodes are used to create a $s$-dimensional node embedding for all real and imaginary parts of the bus voltages, $\Re({V}_i)$ and $\Im({V}_i)$, which are used to generate state variable predictions. Factor nodes, two per measurement phasor, serve as inputs for the measurement values, variances, and covariances, also given in rectangular coordinates. These values are then transformed and sent to variable nodes via GNN message passing. Unlike the approximative WLS SE, the GNN includes measurement covariances in the inputs without increasing the computational complexity. To achieve better representation of node's neighbourhood structure, we perform variable node feature augmentation using binary index encoding. Since variable nodes have no additional input features, this encoding allows the GNN to better capture the relationships between nodes. Compared to one-hot encoding used in \cite{kundacina2022state}, binary index encoding significantly reduces the number of input neurons and trainable parameters, as well as training and inference time. Figure~\ref{toyFactorGraph} illustrates a two-bus power system on the left, with a PMU on the first bus, containing one voltage and one current phasor measurement. The nodes connected by full lines represent the corresponding factor graph on the right. 

Unlike approaches that apply GNNs over the bus-branch power system model, such as in \cite{pagnier2021physicsinformed, TGCN_SE_2021}, the using GNNs over factor-graph-like topology allows for the incorporation of various types and quantities of measurements on both power system buses and branches. The ability to simulate the addition or removal of various measurements can be easily achieved by adding or removing factor nodes from any location in the graph. In contrast, using a GNN over the bus-branch power system model would require allocating a single input vector to each bus that includes all potential measurement data for that bus and its neighbouring branches. This can cause problems, such as having to fill elements in the input vector with zeros when not all possible measurements are available, and making the output sensitive to the order of measurements in the input vector. This can make it difficult to accurately model the system and generate reliable results.

Augmenting the factor graph topology by connecting the variable nodes in the $2$-hop neighbourhood significantly improves the model's prediction quality in unobservable scenarios \cite{kundacina2022state}. This is because the graph should remain connected even when we remove factor nodes to simulate measurement loss. This will allow the messages to be still propagated in the whole $K$-hop neighbourhood of the variable node. In other words, a factor node corresponding to a branch current measurement can be removed while still preserving the physical connection between the power system buses. This requires an additional set of trainable parameters for the variable-to-variable message function. Although the augmented factor graph displayed with both full and dashed lines in Fig.~\ref{toyFactorGraph} is not a factor graph because it is no longer bipartite, we will still refer to the nodes as factor and variable nodes for simplicity.

Since the proposed GNN operates on a heterogeneous graph, we use two different types of GNN layers: one for aggregation in factor nodes, and one for variable nodes. These layers, denoted as $\layerf(\cdot|\theta^{\layerf}): \mathbb {R}^{\textrm{deg}(f)  \cdot s} \mapsto \mathbb {R}^{s}$ and $\layerv(\cdot|\theta^{\layerv}): \mathbb {R}^{\textrm{deg}(v) \cdot s} \mapsto \mathbb {R}^{s}$, have their own sets of trainable parameters  $\theta^{\layerf}$ and $\theta^{\layerv}$, allowing their message, aggregation, and update functions to be learned separately. Additionally, we use different sets of trainable parameters for variable-to-variable and factor-to-variable node messages, $\textrm{Message}\textsuperscript{f$\rightarrow$v}(\cdot|\theta^{\textrm{Message}\textsuperscript{f$\rightarrow$v}}): \mathbb {R}^{2s} \mapsto \mathbb {R}^{u}$ and $\textrm{Message}\textsuperscript{v$\rightarrow$v}(\cdot|\theta^{\textrm{Message}\textsuperscript{v$\rightarrow$v}}): \mathbb {R}^{2s} \mapsto \mathbb {R}^{u}$, in the $\layerv(\cdot|\theta^{\layerv})$ layer. In both GNN layers, we use two-layer feed-forward neural networks as message functions, single layer neural networks as update functions and the attention mechanism in the aggregation function. Furthermore, we apply a two-layer neural network $\pred(\cdot|\theta^{\pred}): \mathbb {R}^{s} \mapsto \mathbb {R}$ to the final node embeddings $\mathbf h^K$ of variable nodes only, to create state variable predictions $\mathbf{x^{pred}}$. For factor and variable nodes with indices $f$ and $v$, neighbourhood aggregation and state variable prediction can be described as: 
\begin{equation}
    \begin{gathered}
        \mathbf {h_v}^k = \layerv(\{\mathbf {h_i}^{k-1} | i \in \mathcal{N}_v\} | \theta^{\layerv})\\
        \mathbf {h_f}^k = \layerf(\{\mathbf {h_i}^{k-1} | i \in \mathcal{N}_f\} | \theta^{\layerf})\\
        {x_v}^{pred} = \pred(\mathbf {h_v}^K|\theta^{\pred})\\
        k \in \{1,\dots,K\},
    \end{gathered}
    \label{embeddingds_and_predictions}
\end{equation}
where $\mathcal{N}_v$ and $\mathcal{N}_f$ denote the $1$-hop neighbourhoods of the nodes $v$ and $f$. All the trainable parameters $\theta$ of the GNN are updated by applying gradient descent, using backpropagation, to a loss function calculated over a mini-batch of graphs. This loss function is the mean-squared difference between the predicted state variables and their corresponding ground-truth values:
\begin{equation} \label{loss_function}
    \begin{gathered}
        L(\theta) = \frac{1}{2nB} \sum_{i=1}^{2nB}({{x_i}^{pred}} - {{x_i}^{label}})^2 \\
        \theta = \{\theta^{\layerv} \mathop{\cup} \theta^{\layerf} \mathop{\cup} \theta^{\pred}\} \\
        \theta^{\layerv} = \{\theta^{\textrm{Message}\textsuperscript{f$\rightarrow$v}} \mathop{\cup} \theta^{\textrm{Message}\textsuperscript{v$\rightarrow$v}} \mathop{\cup} \theta^{\textrm{Aggregate}\textsuperscript{v}} \mathop{\cup} \theta^{\textrm{Update}\textsuperscript{v}}\}\\
        \theta^{\layerf} = \{\theta^{\textrm{Message}\textsuperscript{v$\rightarrow$f}} \mathop{\cup} \theta^{\textrm{Aggregate}\textsuperscript{f}} \mathop{\cup} \theta^{\textrm{Update}\textsuperscript{f}}\},
    \end{gathered}
\end{equation}
where the total number of variable nodes in a graph is $2n$, and the number of graphs in the mini-batch is $B$. 
In this work, we chose the loss function for training the GNN based on the fundamental SE problem, where state variables are obtained from available measurement information. However, if there is a requirement to include additional constraints in the SE calculation, it is possible to achieve this by adding new terms to the loss function defined in \eqref{loss_function}. For example, the loss function can be augmented with the power balance error at each bus where the constraints are imposed, in addition to minimising the prediction error from the labels. A similar approach has been proposed in \cite{LOPEZ2023GNNpowerFlow}, where the power flow problem is solved using GNN by minimising the power balance errors at each bus. Adding additional constraints (e.g., zero injection constraints) to the GNN SE loss function can improve the SE results, especially in distributed power systems with limited measurement coverage.

\begin{figure}[!t]
\centering
\subfloat[]{
\begin{tikzpicture} [scale=0.75, transform shape]

\tikzset{
    varNode/.style={circle,minimum size=6mm,fill=white,draw=black},
    factorVoltage/.style={draw=black,fill=red!80, minimum size=6mm},
    factorCurrent/.style={draw=black,fill=orange!80, minimum size=6mm},
    box/.style={draw, fill=Goldenrod, minimum width=1.5cm, minimum height=0.9cm}}

\begin{scope}[local bounding box=graph]

\node [box]  (layerF0) at (-3, 1.5) {$\layerf$};
\node [box]  (layerV0) at (-3, 0) {$\layerv$};

\node[factorVoltage, label=above:$\mathbf{h_{f}}^{K-1}$] (facReV1) at (0, 1.5) {};
\node[varNode, label=below:$\mathbf{h_{v2}}^{K-1}$] (varImV1) at (0, 0) {};
\node[varNode, label=above:$\mathbf{h_{v}}^{K}$] (varReV1) at (6, 0) {};

\node [box]  (layerV1) at (3, 0) {$\layerv$};
\node [box]  (pred) at (6, -3) {$\pred$};
\node [box, fill=SpringGreen]  (loss) at (3, -3) {Loss};

\draw[-stealth] (-4.25, 1.75) -- (layerF0.west);
\draw[-stealth] (-4.25, 1.5) -- (layerF0.west) node[at start,left]{$...$};
\draw[-stealth] (-4.25, 1.25) -- (layerF0.west);

\draw[-stealth] (-4.25, 0.25) -- (layerV0.west);
\draw[-stealth] (-4.25, 0) -- (layerV0.west) node[at start,left]{$...$};
\draw[-stealth] (-4.25, -0.25) -- (layerV0.west);

\draw[-stealth] (layerF0.east) -- (facReV1.west);
	
\draw[-stealth] (layerV0.east) -- (varImV1.west);
	
\draw[-stealth] (varImV1.east) -- (layerV1.west);
	
\draw[-stealth] (0.25, -2.5) -- (layerV1.west)
	node[at start,left]{$...$} node[very near end,left]{$...$};
	
\draw[-stealth] (facReV1.east) -- (layerV1.west);
	
\draw[-stealth] (layerV1.east) -- (varReV1.west);
	
\draw[-stealth] (varReV1.south) -- (pred.north);
	
\draw[-stealth] (pred.west) -- (loss.east)
	node[midway,above]{$output$};
	
\draw[-stealth] (3, -2.1) -- (loss.north)
	node[at start,above]{$label$};

\end{scope}

\end{tikzpicture}
}
\hfil
\subfloat[]{

\begin{tikzpicture} [scale=0.68, transform shape]

\tikzset{
    box/.style={draw, fill=Goldenrod, minimum width=1.5cm, minimum height=0.8cm}}
    
\begin{scope}[local bounding box=graph]

\node [box]  (message1) at (0, 1.5) {Message\textsuperscript{f$\rightarrow$v}};
\node [box]  (message2) at (0, 0) {Message\textsuperscript{v$\rightarrow$v}};
\node [box]  (gat) at (3, 0) {\textrm{GAT}\textsuperscript{v}};
\node [box]  (update) at (5.5, 0) {\textrm{Update}\textsuperscript{v}};


\draw[-stealth] (-1.8, 0) -- (message2.west) node[at start,left]{$\mathbf{h_{v2}}^{K-1}; \mathbf{h_{v}}^{K-1}$};

\draw[-stealth] (-1.8, 1.5) -- (message1.west) node[at start,left]{$\mathbf{h_{f}}^{K-1}; \mathbf{h_{v}}^{K-1}$};

\draw[-stealth] (message1.east) -- (gat.west);

	
\draw[-stealth] (message2.east) -- (gat.west);
	
\draw[-stealth] (2, -1) -- (gat.west)
	node[at start,left]{$...$} node[very near end,left]{$...$};
	
\draw[-stealth] (gat.east) -- (update.west);

\draw[-stealth] (update.east) -- (7.5, 0) node[at end,right]{$\mathbf{h_{v}}^{K}$};

\draw[-stealth] (5.5, -1.0) -- (update.south) node[at start,below]{$\mathbf{h_{v}}^{K-1}$};

\end{scope}

\label{layerDetails}

\end{tikzpicture}
}
\caption{Subfigure (a) shows a high-level computational graph that starts with the loss function for the output of a variable node $v$. Subfigure (b) depicts the detailed structure of a single GNN $\layerv$. Functions with trainable parameters are highlighted in yellow.}
    \label{computationalGraph}
\end{figure}
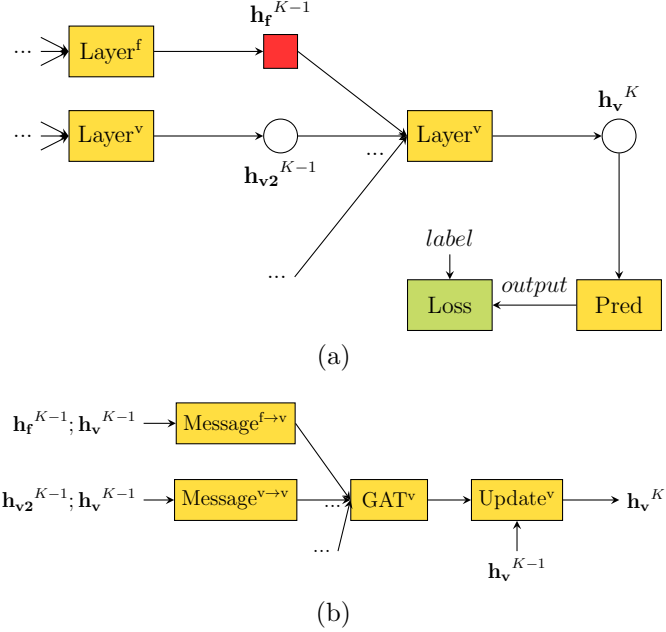

Fig.~\ref{computationalGraph} shows the high-level computational graph for the output of a variable node from the augmented factor graph given in Fig.~\ref{toyFactorGraph}. For simplicity, only one unrolling of the neighbourhood aggregation is shown, as well as only the details of the parameters $\theta^{\layerv}$.

\subsection{Computational Complexity and Distributed Inference} 
Because the node degree in the power system graph does not increase with the total number of nodes, the same is true for the node degrees in the augmented factor graph. This means that the inference time per variable node remains constant, as it only requires information from the node's $K$-hop neighbourhood, whose size also does not increase with the total number of nodes. This implies that the computational complexity of inferring all state variables is $\mathcal{O}(n)$. To avoid the over-smoothing problem in GNNs \cite{Chen_2020_oversmoothing}, a small value is assigned to $K$, thus not affecting the overall computational complexity of the inference.

To make the best use of the proposed approach for large-scale power systems, the inference should be performed in a computationally and geographically distributed manner. This is necessary because the communication delays between the PMUs and the central processing unit can hinder the full utilisation of the PMUs' high sampling rates. The distributed implementation is possible as long as all the measurements within a node's $K$-hop neighbourhood in the augmented factor graph are fed into the computational module that generates the predictions. For any arbitrary $K$, the GNN inference method only requires PMUs that are physically located within the $\lceil K/2 \rceil$-hop neighbourhood of the power system bus.

\section{Results}
In this section, we describe the training setups used for augmented factor graph-based GNN models and assess the prediction quality of the trained models in various test scenarios. Training, validation, and test sets are obtained using WLS solutions of the system described in (\ref{SE_system_of_lin_eq}) to label various samples of input measurements. The measurements are obtained by adding Gaussian noise to the exact power flow solutions, with each calculation performed using a different, randomly sampled load profile to capture a wide range of power system states. Due to the strong interpolation and extrapolation abilities of GNNs \cite{xu2021how}, our method of randomly sampling from a wide diversity of loads for training examples is effective for generalising the GNN algorithm for state estimation under varying load conditions. GNN models are tested in three different situations: i) optimal number of PMUs (minimal measurement redundancy, for which the WLS SE offers a solution); ii) underdetermined scenarios; iii) scenarios with maximal measurement redundancy. Maximal measurement redundancy refers to the highest possible value of measurement redundancy, which is defined as the ratio $\frac{m}{n}$, where $m$ is the number of phasor measurements, and $n$ is the number of buses in the power system. In other words, it represents the situation where a phasor measurement is installed at every possible location in the power system once, which includes complex voltage measurements on all the buses and complex current measurements on all the branch ends. We also compare the proposed approach with the state-of-the-art deep neural network (DNN)-based SE algorithms and assess its scalability, sample efficiency, and robustness to outliers. The hyperparameters used throughout all the experiments include a node embedding size of $s=64$, a learning rate of $4\times 10^{-4}$, a minibatch size of $B=32$, $K=4$ GNN layers, ReLU activation functions, and Adam optimiser\footnote{Source code is available online at https://github.com/ognjenkundacina.}. All the results presented in this paper are normalised using the corresponding nominal voltages in the test power systems and a base power of 100 MVA.


\subsection{Power System With Optimally Placed PMUs}
\label{subsec:optPlacedPMUs}

In this subsection, we conduct a series of experiments on the IEEE 30-bus power system, using measurement variances of $10^{-5}$, $10^{-3}$, and $10^{-1}$ for the creation of the training, validation, and test sets. We used higher variance values to model misalignments in phasor measurements caused by communication delays \cite{zhao2019power}. The number and positions of PMUs are fixed and determined using the optimal PMU placement algorithm \cite{optimalPMUPlacement}, which finds the smallest set of PMUs that make the system observable. This algorithm has resulted in a total of $10$ PMUs and $50$ measurement phasors, $10$ of which are voltage phasors and the rest are current phasors.

Table \ref{tbl_MSEsIEEE30} shows the $100$-sample test set results for all the experiments on the IEEE 30-bus power system, in the form of average mean square errors (MSEs) between the GNN predictions and the test set labels. These results are compared with the average MSE between the labels and the approximate WLS SE solutions defined in Sec. \ref{sec:lse}\footnote{It is important to note that the variance of the input measurement and the MSE, which measures the difference between predicted and actual values to assess the quality of state estimates, are not directly connected. The variance in measurements corresponds to the additional noise introduced to voltage and current measurement phasors. Meanwhile, MSEs in Table \ref{tbl_MSEsIEEE30} are computed by comparing the GNN's predictions of bus voltages, as well as the approximate WLS SE results, to the actual results of the WLS solutions of the noisy system of linear equations. However, it should be pointed out that there is a trend where larger variances lead to larger MSE values.}. The results show that for systems with optimally placed PMUs and low measurement variances, GNN predictions have very small deviations from the exact WLS SE, although they are outperformed by the approximate WLS SE. For higher measurement variances, GNN has a lower estimation error than the approximate WLS SE, while also having lower computational complexity in all cases.

\begin{table}[htbp]
\caption{Comparison of GNN and approximative SE test set MSEs for various measurement variances.}
\begin{center}
    \begin{tabular}{ | c | c | c | }
        \hline
        \textbf{Variances} & \textbf{GNN}  & \textbf{Approx. SE} \\
        \hline
        $10^{-5}$ & $2.48\times 10^{-6}$ & $1.87\times 10^{-8}$ \\
        \hline
        $10^{-3}$ & $8.21\times 10^{-6}$ & $2.25\times 10^{-6}$ \\
        \hline
        $10^{-1}$ & $7.47\times 10^{-4}$ & $2.27\times 10^{-3}$ \\
        \hline
    \end{tabular}
\label{tbl_MSEsIEEE30}
\end{center}
\end{table}

\begin{figure}[htbp]
    \centering
    \pgfplotstableread[col sep = comma,]{data/PredictionsPerNode0Excluded.csv}\CSVPredVSObservedZeroExcludedEdgesVarNodes
    
        \begin{tikzpicture}
        
        \begin{groupplot}[group style={group size=1 by 2, vertical sep=0.5cm}, ]
        \nextgroupplot[
        yscale=0.55,
        legend cell align={left},
        legend columns=1,
        legend style={
          fill opacity=0.8,
          font=\small,
          draw opacity=1,
          text opacity=1,
          at={(0.75,1.1)},
          anchor=south,
          draw=white!80!black
        },
        scaled x ticks=manual:{}{\pgfmathparse{#1}},
        tick align=outside,
        tick pos=left,
        x grid style={white!69.0196078431373!black},
        xmin=-1.45, xmax=30.45,
        xtick style={color=black},
        xticklabels={},
        y grid style={white!69.0196078431373!black},
        ylabel={\(\displaystyle \Re({V}_i)\) [p.u.]},
        ymin=0.93, ymax=1.08,
        ytick style={color=black},
        xmajorgrids=true,
        ymajorgrids=true,
        grid style=dashed
        ]
        \addplot [red, mark=square*, mark options={scale=0.5}, mark repeat=2,mark phase=1]
        table [x expr=\coordindex, y={predictionsRE}]{\CSVPredVSObservedZeroExcludedEdgesVarNodes};
        \addlegendentry{Predictions}
        
        \addplot [blue, mark=square*, mark options={scale=0.5}, mark repeat=2,mark phase=2]
        table [x expr=\coordindex, y={true_valuesRE}]{\CSVPredVSObservedZeroExcludedEdgesVarNodes};
        \addlegendentry{Labels}
        
        
        
        
        \nextgroupplot[
        yscale=0.55,
        tick align=outside,
        tick pos=left,
        x grid style={white!69.0196078431373!black},
        xlabel={Bus index ($i$)},
        xlabel style={yshift=-1pt},
        xmin=-1.45, xmax=30.45,
        xtick style={color=black},
        y grid style={white!69.0196078431373!black},
        ylabel={\(\displaystyle \Im({V}_i)\) [p.u.]},
        ymin=-0.35, ymax=0.05,
        ytick style={color=black},
        xmajorgrids=true,
        ymajorgrids=true,
        grid style=dashed
        ]
        
        \addplot [red, mark=square*, mark options={scale=0.5}, mark repeat=2,mark phase=1]
        table [x expr=\coordindex, y={predictionsIM}]{\CSVPredVSObservedZeroExcludedEdgesVarNodes};
        
        \addplot [blue, mark=triangle*, mark options={scale=0.9}, mark repeat=2,mark phase=2]
        table [x expr=\coordindex, y={true_valuesIM}]{\CSVPredVSObservedZeroExcludedEdgesVarNodes};
        
                
        
        
        \end{groupplot}
        
        \end{tikzpicture}
    \caption{GNN predictions and labels for one test example with optimally placed PMUs.}
    \label{PredictionsPerNode0Excluded}
\end{figure}
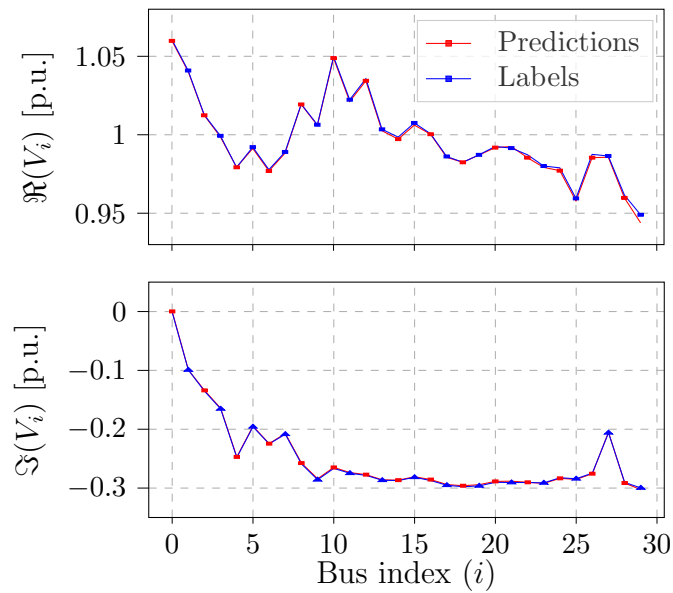

In Fig.~\ref{PredictionsPerNode0Excluded}, we present the predictions and labels for each of the variable nodes for one of the samples from the test set created with measurement variance $10^{-5}$. The results for the real and imaginary parts of the complex node voltages (shown in the upper and lower plots, respectively) indicate that GNNs can be used as accurate SE approximations.

\subsection{Performance in a Partially Observable Scenario} \label{subsecPartiallyObservable}
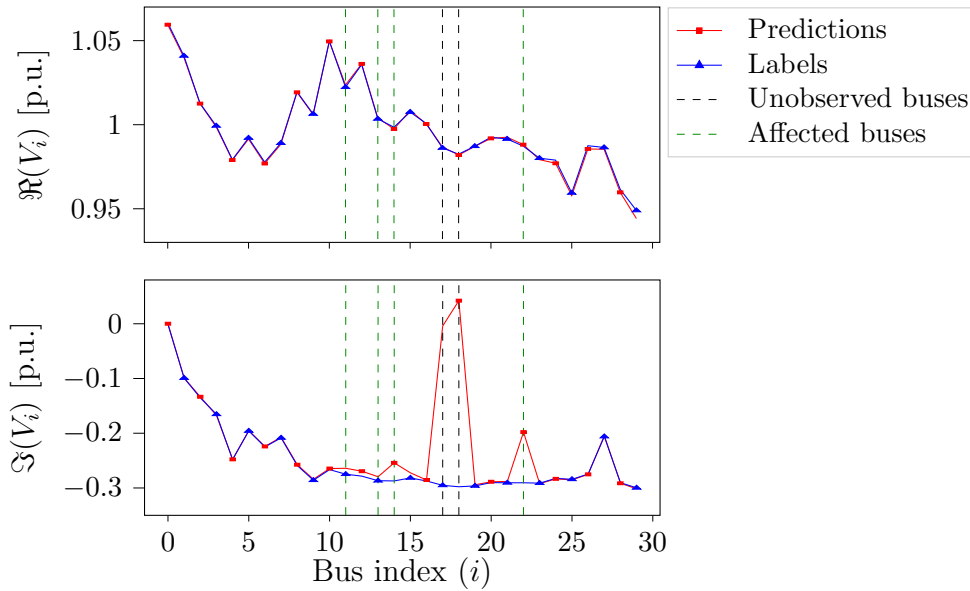
\begin{figure}[htbp]
    \centering
    \pgfplotstableread[col sep = comma,]{data/PredictionsPerNode2PMUExcluded.csv}\CSVPredVSObservdTwoPMUExcludedEdgesVarNodes
    
        \begin{tikzpicture}
        
        \begin{groupplot}[group style={group size=1 by 2, vertical sep=0.5cm}, ]
        \nextgroupplot[
        yscale=0.55,
        legend cell align={left},
        legend columns=1,
        legend style={
          fill opacity=1,
          font=\small,
          draw opacity=1,
          text opacity=1,
          at={(1.32,17.78)},
          anchor=south,
          draw=white!80!black,
          nodes={scale=0.98, transform shape}
        },
        scaled x ticks=manual:{}{\pgfmathparse{#1}},
        tick align=outside,
        tick pos=left,
        x grid style={white!69.0196078431373!black},
        xmin=-1.45, xmax=30.45,
        xtick style={color=black},
        xticklabels={},
        y grid style={white!69.0196078431373!black},
        ylabel={\(\displaystyle \Re({V}_i)\) [p.u.]},
        ymin=0.93, ymax=1.07,
        ytick style={color=black}
        ]
        \addplot [red, mark=square*, mark options={scale=0.5}, mark repeat=2,mark phase=1]
        table [x expr=\coordindex, y={predictionsRE}]{\CSVPredVSObservdTwoPMUExcludedEdgesVarNodes};
        \addlegendentry{Predictions}
        
        \addplot [blue, mark=triangle*, mark options={scale=0.9}, mark repeat=2,mark phase=2]
        table [x expr=\coordindex, y={true_valuesRE}]{\CSVPredVSObservdTwoPMUExcludedEdgesVarNodes};
        \addlegendentry{Labels}

        

        \addplot [very thin, black, dashed]
        table {%
        17 -2.0
        17 2.0
        };
        \addlegendentry{Unobserved buses}

        \addplot [very thin, green!55!black, dashed]
        table {%
        14 -2
        14 2
        };
        \addlegendentry{Affected buses}

        \addplot [very thin, green!55!black, dashed, forget plot]
        table {%
        11 -2.0
        11 2.0
        };
        \addplot [very thin, green!55!black, dashed, forget plot]
        table {%
        13 -2.0
        13 2.0
        };
        \addplot [very thin, green!55!black, dashed, forget plot]
        table {%
        22 -2.0
        22 2.0
        };

        \addplot [very thin, black, dashed, forget plot]
        table {%
        18 -2.0
        18 2.0
        };

        \nextgroupplot[
        yscale=0.55,
        legend cell align={left},
        legend columns=1,
        legend style={
          fill opacity=0.8,
          font=\small,
          draw opacity=1,
          text opacity=1,
          at={(0.785,4.249)},
          anchor=south,
          draw=white!80!black,
          nodes={scale=0.98, transform shape}
        },
        tick align=outside,
        tick pos=left,
        x grid style={white!69.0196078431373!black},
        xlabel={Bus index ($i$)},
        xlabel style={yshift=-1pt},
        xmin=-1.45, xmax=30.45,
        xtick style={color=black},
        y grid style={white!69.0196078431373!black},
        ylabel={\(\displaystyle \Im({V}_i)\) [p.u.]},
        ymin=-0.35, ymax=0.08,
        ytick style={color=black}
        ]
        
        \addplot [red, mark=square*, mark options={scale=0.5}, mark repeat=2,mark phase=1, forget plot]
        table [x expr=\coordindex, y={predictionsIM}]{\CSVPredVSObservdTwoPMUExcludedEdgesVarNodes};
        
        \addplot [blue, mark=triangle*, mark options={scale=0.9}, mark repeat=2,mark phase=2, forget plot]
        table [x expr=\coordindex, y={true_valuesIM}]{\CSVPredVSObservdTwoPMUExcludedEdgesVarNodes};
        
        
        
        \addplot [very thin, green!55!black, dashed, forget plot]
        table {%
        14 -2
        14 2
        };
        
        \addplot [very thin, green!55!black, dashed, forget plot]
        table {%
        11 -2.0
        11 2.0
        };
        \addplot [very thin, green!55!black, dashed, forget plot]
        table {%
        13 -2.0
        13 2.0
        };
        \addplot [very thin, green!55!black, dashed, forget plot]
        table {%
        22 -2.0
        22 2.0
        };
        
        \addplot [very thin, black, dashed, forget plot]
        table {%
        17 -2.0
        17 2.0
        };
        
        \addplot [very thin, black, dashed, forget plot]
        table {%
        18 -2.0
        18 2.0
        };
        
        \end{groupplot}
        
        \end{tikzpicture}

    \caption{GNN predictions and labels for one test example with phasors from two neighbouring PMUs removed. Vertical black lines indicate unobserved buses, while green lines represent buses that are affected by the loss of measurement data.}
    \label{PredictionsPerNode2PMUExcluded}
\end{figure}

To further assess the robustness of the proposed model, we test it by excluding several measurement phasors from the previously used test samples with optimally placed PMUs, resulting in an underdetermined system of equations that describes the SE problem. These scenarios are relevant even at higher levels of system redundancy, where partial grid observability can occur due to multiple component (PMU and communication link) failures caused by natural disasters or cyberattacks. To exclude a measurement phasor from the test sample, we remove its real and imaginary parts from the input data, which is equivalent to removing two factor nodes from the augmented factor graph. We use the previously used 100-sample test set to create a new test set by removing selected measurement phasors from each sample while preserving the same labels obtained as SE solutions of the system with all the measurements present. As an example, we consider a scenario where two neighbouring PMUs fail to deliver measurements to the state estimator. In this case, all eight measurement phasors associated with the removed PMUs are excluded from the GNN inputs. The average MSE for the test set of 100 samples created by removing these measurements from the original test set used in this section is $3.45\times 10^{-3}$. The predictions and labels for a single test set sample, per variable node index, are shown in Figure \ref{PredictionsPerNode2PMUExcluded}. The figure includes vertical black dashed lines that indicate the indices of unobserved buses 17 and 18. These buses have higher prediction errors due to the lack of input measurement data. Neighbouring buses that are not unobserved, but are affected by measurement loss, are indicated with vertical green lines and have lower prediction errors. It can be observed that significant deviations from the labels occur for some of the neighbouring buses, while the GNN predictions are a decent fit for the remaining node labels. This demonstrates the proposed model's ability to sustain error in the neighbourhood of the malfunctioning PMU, as well as its robustness in scenarios that cannot be solved using standard WLS approaches. A possible explanation for the higher susceptibility to errors in the imaginary parts of the voltage is related to their variance in the training set. The variance of real parts of voltages is $6.6 \times 10^{-4}$, while the variance of imaginary parts of voltages is $4.8 \times 10^{-3}$. This indicates that the imaginary parts of voltages have a higher variability and may therefore be more prone to errors in the prediction model.

\subsection{Comparison With the Feed-Forward Deep Neural Network-Based SE}
\label{subsec:compareWithDNN}

The main goal of this subsection is to compare the performance of the proposed GNN-based SE approach with a state-of-the-art deep learning-based approach on a variety of power systems. We used a 6-layer feed-forward DNN model, proposed by \cite{zhang2019}, with the same number of neurons in each layer as the number of input measurement scalars. This DNN architecture has similar performance as the best architecture proposed in the same work obtained by unrolling the iterative nonlinear SE solver, which cannot be applied directly to the linear SE problem we are considering.

Secondarily, this subsection aims to evaluate the scalability of the proposed GNN. This will be achieved by analysing its performance on power systems of varying sizes and benchmarking it against the mentioned state-of-the-art in deep learning-based SE. We tested both approaches on the IEEE 30-bus, IEEE 118-bus, IEEE 300-bus, and the ACTIVSg 2000-bus power systems \cite{ACTIVSg2000}, with measurement variances set to $10^{-5}$. In contrast to previous examples, we used maximal measurement redundancies, ranging from $3.73$ to $4.21$. We provide a comparison of the number of trainable parameters for both GNN and DNN models for various power system sizes, which is often left out in similar analyses. To compare sample efficiencies between the GNN and DNN approaches, separate models for each of the test power systems were trained using smaller and larger datasets, containing 10 and 10000 samples, except for the 2000-bus power system, where a 1000-sample training set was used\footnote{Training the model for the ACTIVSg 2000-bus power systems on $10000$ samples could not be performed due to requirements that exceeded our hardware configuration (Intel Core(TM) i7-11800H 2.30 GHz CPU, and 16 GB of RAM). Optimising the batch loading of the training examples would make the training process feasible for large datasets even on a modest hardware setup.}. The results for all test power systems are presented in Table \ref{tbl_GNN_vs_DNN}. The first four rows show the 100-sample test set MSE for GNN and DNN models trained using smaller and larger datasets. The last two rows of the table show the number of trainable parameters for both approaches, depending on the power system size.

\begin{table}[]
\caption{A comparison of the performance of GNN and DNN models trained on different training set sizes, as measured by test set MSE and the number of trainable parameters.}
\begin{center}
\resizebox{13.7cm}{!}{
\begin{tabular}{|c|c|c|c|}
\hline
\textbf{}  & \textbf{IEEE 30}                                          &  \textbf{IEEE 300} & \textbf{ACTIVSg 2000} \\ \hline
Small training set GNN    & \boldmath{$4.73 \times 10^{-6}$} & \boldmath{$5.94\times 10^{-5}$} & \boldmath{$5.08\times 10^{-4}$} \\ \hline
Small training set DNN  & $9.29\times 10^{-4}$ & $5.92\times 10^{-3}$ & $4.77\times 10^{-3}$ \\ \Xhline{5\arrayrulewidth}
Large training set GNN  & \boldmath{$2.48\times 10^{-6}$} & \boldmath{$6.62\times 10^{-6}$} & \boldmath{$3.91\times 10^{-4}$} \\ \hline
Large training set DNN & $6.28\times 10^{-6}$ & $2.91\times 10^{-3}$ & $2.61\times 10^{-3}$ \\ \Xhline{5\arrayrulewidth} 

GNN parameters & \boldmath{$4.99\times 10^{4}$} & \boldmath{$4.99\times 10^{4}$} & \boldmath{$4.99\times 10^{4}$}   \\ \hline
DNN parameters & $3.16\times 10^{5}$ & $3.15\times 10^{7}$ & {$1.77\times 10^{9}$}   \\ \hline
\end{tabular}
\label{tbl_GNN_vs_DNN}
}
\end{center}
\end{table}

The results show that the GNN approaches result in \textbf{higher overall accuracy} compared to the corresponding DNN approaches for all the power system and training set sizes. Furthermore, the number of trainable parameters (i.e., the model size) is constant\footnote{More precisely, the number of trainable parameters in the proposed GNN model remains nearly constant as the number of buses in the power system increases. This effect would only be noticeable for larger power systems. The only exception is the number of input neurons for the binary index encoding of the variable nodes, which grows logarithmically with the number of variable nodes. However, this increase is insignificant compared to the total number of GNN parameters.} and relatively low for GNN models, because the number of neurons in a layer is constant regardless of power system size. In contrast, the number of parameters grows quadratically for DNN models, because the number of neurons in a layer grows linearly with the input size, resulting in quadratic growth of the trainable parameter matrices. When expressed in computer memory units, the GNN models we used had a significantly smaller memory footprint, taking up only 0.19 MB. In comparison, the DNN model used for the ACTIVSg 2000 power system required a much larger 6.58 GB of memory, resulting in more challenging training and inference processes. The high number of trainable parameters required by DNN models increases their storage requirements, increases the dimensionality of the training process, and directly affects the \textbf{inference speed and computational complexity}. Since the proposed GNNs have a linear computational complexity in the prediction process, one training iteration of GNN also has a linear computational complexity. In contrast, one training iteration of DNN-based SE would have at least quadratic computational complexity per training iteration, making the overall training process significantly slower. To recall, the reason why the GNN has a constant number of parameters and generates predictions with linear computational complexity is that it takes measurements from a limited $K$-hop neighbourhood for every node, regardless of the size of the power system.

The results indicate that the quality of GNN and DNN model predictions improves with more training data. However, compared to GNNs, DNN models performed significantly worse on smaller datasets, suggesting that they are \textbf{less sample efficient} and more prone to overfitting due to their larger number of parameters. While we used randomly generated training sets in the experiments, narrowing the learning space by selecting training samples based on historical load consumption data could potentially result in even better performance with small datasets.

The use of GNNs for power systems analysis has several additional advantages over using DNNs. One advantage is flexibility: spatial GNNs can produce results even if the input power system topology changes, whereas conventional DNN methods are trained and tested on the same topology of the power system. For example, if some measurements are removed from the inputs (as discussed in Subsection \ref{subsecPartiallyObservable}), a DNN would require retraining from scratch with the new topology. GNNs also have some theoretical advantages over other deep learning methods in that they are permutation invariant and equivariant. This means that the output of the GNN is the same regardless of the order in which the inputs are presented, and that the GNN output changes in a predictable and consistent way when the inputs are transformed. This property is useful for problems like SE, where the order of the nodes and edges is not important and the system can undergo topological changes. In addition, GNNs incorporate topology information into the learning process by design, whereas many other deep learning methods in power systems use node-level data as inputs while ignoring connectivity information. Finally, unlike most deep learning methods, spatial GNNs can be distributed for evaluation among edge devices.

Similar conclusions can be drawn when comparing GNNs with recurrent and convolutional neural networks for similar power systems' analysis algorithms, as they also require information from the entire power system as input. Overall, this comparison highlights the potential advantages of using GNNs for power system modelling and analysis.

\subsection{Robustness to Outliers}
To assess how well the proposed model can handle outliers in input data, we carried out experiments on two separate test sets, each containing samples with different degrees of outlier intensity. The experiments followed the setup described in Sec. \ref{subsec:optPlacedPMUs}, with the test samples initially generated using a measurement variance of $10^{-5}$. We replaced one of the existing measurements in each test sample with a randomly generated value, using a variance of either $\sigma_1 = 1.6$ or $\sigma_2 = 1.6\cdot10^2$. WLS SE solutions without outliers in the inputs are used as ground-truth values. The results, shown in Table \ref{tbl_outliers}, indicate the performance of four different approaches, as well as the WLS SE and the approximative WLS SE algorithm on the same test sets.

The first approach, which uses the already trained model from Sec. \ref{subsec:optPlacedPMUs}, results in the highest prediction error on tests set with outliers. The primary factor contributing to the MSE, particularly in the test set with outliers generated using larger variances, are significant mismatches from the ground-truth values in the $K$-hop neighbourhood of the outlier. This occurs because the ReLU activation function does not constrain its inputs during neighbourhood aggregation. To address this problem, we propose a second approach where we train a GNN model with the same architecture as the previous one, but which uses the tanh (hyperbolic tangent) activation function instead of ReLU. As presented in Table \ref{tbl_outliers}, this approach results in a significantly lower test set MSE for outliers generated using larger variances compared to the proposed GNN with ReLU activations, WLS SE, and the approximative WLS SE. The saturation effect of tanh prevents high values stemming from outliers from propagating through the GNN, but also reduces the training quality due to the vanishing gradient problem. Specifically, all the experiments we conducted under the same conditions for GNN with tanh activations required more epochs to converge to a solution with lower prediction quality compared to the GNN with ReLU activations. As a third approach, we propose training a GNN model with ReLU activations on a dataset in which half of the samples contain outliers, which are generated in the same manner as the test samples used in this subsection. This approach turned out to be the most effective for both test cases because the GNN learns to neutralise the effect of unexpected inputs from the dataset examples while maintaining accurate predictions in the absence of outliers in the input data. To confirm the validity of this methodology, we trained the DNN introduced in the subsection \ref{subsec:compareWithDNN} using the same datasets containing outliers. DNN was able to neutralise the effect of unexpected inputs because the input power system is small, resulting in the second-best approach in terms of robustness to outliers, trailing the GNN trained on the dataset containing outliers. 

In summary, as expected, all methods produced better results for the test set containing outliers with lower variances, while the GNN trained with outliers demonstrated the best performance for both higher and lower variance outliers. We note that these are only preliminary efforts to make the GNN model robust to outliers, and that future research could combine ideas from standard bad data processing methods in SE with the proposed GNN approach.

\begin{table}[htbp]
\caption{  A comparison of the results of various approaches for two test sets with different degrees of outlier intensity.}
\begin{center}
    \begin{tabular}{ | c | c | c | }
        \hline
        \textbf{Approach} & \begin{tabular}[c]{@{}c@{}}\textbf{Test set MSE} \\ $\mathbf{\sigma_1 = 1.6}$\end{tabular}  & \begin{tabular}[c]{@{}c@{}}\textbf{Test set MSE} \\ $\mathbf{\sigma_2 = 1.6 \cdot 10^2}$\end{tabular} \\
        \hline
        GNN & $9.48\times 10^{-3}$ & $1.60\times 10^{3}$ \\
        \hline
        GNN with tanh & $6.87\times 10^{-3}$ & $2.39\times 10^{-2}$ \\
        \hline
        GNN trained with outliers & \boldmath{$4.44\times 10^{-6}$} & \boldmath{$7.99\times 10^{-6}$} \\
        \hline
        DNN trained with outliers & $1.06\times 10^{-5}$ & $4.21\times 10^{-5}$ \\
        \hline
        WLS SE & $1.43\times 10^{-3}$ & $1.41\times 10^{-1}$ \\
        \hline
        Approximative WLS SE & $1.39\times 10^{-3}$ & $1.35\times 10^{-1}$ \\
        \hline
    \end{tabular}
\label{tbl_outliers}
\end{center}
\end{table}

\section{Conclusions}
In this research, we investigate how GNN can be used as fast solver of linear SE with PMUs. The proposed graph attention network-based model, specialised for the newly introduced heterogeneous augmented factor graphs, recursively propagates the input measurements from the factor nodes to generate predictions in the variable nodes. Evaluating the trained model on unseen data samples confirms that the proposed GNN approach can be used as a highly accurate approximator of the linear WLS SE solutions, with the added benefit of linear computational complexity at inference time. The model is robust in unobservable scenarios that are not solvable using standard WLS SE and deep learning methods, such as when individual phasor measurements or entire PMUs fail to deliver measurement data to the proposed SE solver. Furthermore, when measurement variances are high or outliers are present in the input data, the GNN model outperforms the approximate WLS SE. The proposed approach demonstrates scalability and sample efficiency when tested on power systems of various sizes, as it makes good predictions even when trained on a small number of randomly generated samples. Finally, the proposed GNN model outperforms the state-of-the-art deep learning-based SE approach in terms of prediction accuracy and significantly lower number of trainable parameters, especially as the size of the power system grows.

In this work, we focused on using GNNs to solve a linear SE model with only phasor measurements. However, the proposed learning framework, graph augmentation techniques, and conclusions can be applied to a wide range of SE formulations. For example, our ongoing research employs the proposed framework for the nonlinear SE model, which includes both PMUs and legacy measurements provided by the supervisory control and data acquisition system, with preliminary results available in \cite{kundacina2022NonlinearSE}. Additionally, the GNN's ability to provide relevant solutions in underdetermined scenarios suggests that it could be useful for GNN-based SE in highly unobservable distribution systems.

While our work shows promising results, an important limitation is the inability to quantify the uncertainty of the GNN predictions. However, we are encouraged by ongoing research efforts to address this issue, as quantifying uncertainty for GNN regression remains an open problem. For instance, \cite{MUNIKOTI20231} proposes a Bayesian framework that uses assumed density filtering to quantify aleatoric uncertainty and Monte Carlo dropout to capture epistemic uncertainty in GNN predictions. In light of this, we believe implementing a similar approach represents a promising future research direction.

\section{Acknowledgment}
This paper has received funding from the European Union's Horizon 2020 research and innovation programme under Grant Agreement number 856967.

 \bibliographystyle{elsarticle-num} 
 \bibliography{cas-refs}





\end{document}